\documentclass[%
reprint,
superscriptaddress,
 amsmath,amssymb,
 aps,
 prl,
twocolumn
]{revtex4}

\usepackage{graphicx}% Include figure files
\usepackage{dcolumn}% Align table columns on decimal point
\usepackage{bm}% bold math
\usepackage{mathtools}
\usepackage{dsfont}
\begin{document}

\preprint{APS/123-QED}

\title{Finding symmetry-breaking Order Parameters with Euclidean Neural Networks}% Force line breaks with \\
% \title{Learning Missing Data Implied by Symmetry with Euclidean Neural Networks}% Force line breaks with \\

\author{Tess E. Smidt}
\email{tsmidt@lbl.gov}
\affiliation{
 Computational Research Division, Lawrence Berkeley National Laboratory, Berkeley, CA 94720
}
\affiliation{Center for Advanced Mathematics for Energy Research Applications (CAMERA),  Lawrence Berkeley National Laboratory, Berkeley, CA 94720}

\author{Mario Geiger}
\affiliation{
 Computational Research Division, Lawrence Berkeley National Laboratory, Berkeley, CA 94720
}
\affiliation{
Department of Physics, \'{E}cole polytechnique f\'{e}d\'{e}rale de Lausanne, Lausanne, Switzerland
}%

\author{Benjamin Kurt Miller}
\affiliation{
 Computational Research Division, Lawrence Berkeley National Laboratory, Berkeley, CA 94720
}
\affiliation{Center for Advanced Mathematics for Energy Research Applications (CAMERA),  Lawrence Berkeley National Laboratory, Berkeley, CA 94720}
\affiliation{%
 Department of Mathematics, Freie Universit\"{a}t Berlin, Berlin, Germany
}%

\date{\today}% It is always \today, today,

\begin{abstract}
% Must be <600 characters (currently 586)
Curie's principle states that ``when effects show certain asymmetry, this asymmetry must be found in the causes that gave rise to them''. We demonstrate that symmetry equivariant neural networks uphold Curie's principle and can be used to articulate many symmetry-relevant scientific questions into simple optimization problems. We prove these properties mathematically and demonstrate them numerically by training a Euclidean symmetry equivariant neural network to learn symmetry-breaking input to deform a square into a rectangle and to generate octahedra tilting patterns in perovskites.
\end{abstract}

\maketitle

%\tableofcontents

% \section{Introduction}
Machine learning techniques such as neural networks are data-driven methods for building models that have been successfully applied to many areas of physics, 
% particularly fields that employ computationally expensive calculations and simulations,
such as quantum matter, particle physics, and cosmology \cite{ RevModPhys.91.045002, MLMeetsQuantum}.

All machine learning models can be abstracted as a function $f$ parameterized by learnable weights $W \in \mathbb{R}^N$ that maps vector space $V_1$ to another vector space $V_2$, i.e. $f: V_1 \times W \rightarrow V_2$. Weights are updated by using a loss function which evaluates the performance of the model. In the case of neural networks which are differentiable models, the weights $w\in W$ are updated using the gradients of the loss $\mathcal{L}$ with respect to $w$, $w = w - \eta \frac{\partial \mathcal{L}}{\partial w}$ where $\eta$ is the learning rate.

An important consideration for enhancing the performance and interpretability of these ``black box'' models when used for physics is how to incorporate axioms of symmetry \cite{PhysRevLett.98.146401, PhysRevB.87.184115, schnet, tfn, kondor2018clebsch, 3dsteerable, anderson2019cormorant, PhysRevLett.120.036002}. Building symmetry into the model prevents the model from learning unphysical bias and can lead to new capabilities for investigating physical systems. 

Symmetry invariant models only operate on invariant quantities i.e. scalars, while symmetry equivariant models can preserve equivariant transformations e.g. a change of coordinate system. A function is equivariant under symmetry group $G$ if and only if the group action commutes with the function, i.e. for the group representation $D_1$ and $D_2$ acting on vector space $V_1$ and $V_2$, respectively, $f(D_1(g)x) = D_2(g)f(x), \forall x \in V_1$ and $\forall g \in G$.
While equivariance is more general, invariant models are easier to build; most present-day symmetry-aware models are invariant. However, only symmetry equivariant models can fully express the richness of symmetry-related phenomena of physical systems, e.g. degeneracy and symmetry breaking.

Identifying sources of symmetry breaking is an essential technique for understanding complex physical systems. Many discoveries in physics have been made when symmetry implied something was missing (e.g. the first postulation of the neutrino by Pauli \cite{Brown1978}); many physical phenomena are now understood to be consequences of symmetry breaking \cite{Anderson393}: the mechanism that generates mass \cite{PhysRevLett.13.321, PhysRevLett.13.508, PhysRevLett.13.585}, superconductivity \cite{bcs,PhysRev.117.648}, and phase transitions leading to ferroelectricity \cite{Landau:1937obd}.

In this Letter, we show how symmetry equivariant models can perform symmetry-related tasks without the conventional tools of symmetry analysis (e.g. character tables and related subgroup conventions). Using these networks, we can pose symmetry-related scientific questions as simple optimization problems without using explicit knowledge of the subgroup symmetry of the input or output. These networks can e.g. identify when data (input and output) are not compatible by symmetry, recover missing symmetry-breaking information, find symmetry-intermediate solutions between a given input and target output, and build symmetry-compatible models from limited data. 

These applications are possible due to two properties of symmetry equivariant neural networks that we prove in this Letter: (1) Symmetry equivariant functions exhibit Curie's Principle \cite{Curie1894, chalmers1970curie}; (2) Gradients of an invariant loss function acting on both the network and target outputs can be used to recover the form (representation) of symmetry-breaking information missing from the network input.

We organize this Letter as follows: First, we provide background on symmetry equivariant neural networks. Second, we prove the symmetry properties of the output and gradients of Euclidean symmetry equivariant neural networks and demonstrate them numerically by training a Euclidean neural network to deform a square into a rectangle. Third, we use this technique on a more complex physical example, octahedral tilting in perovskites.

Euclidean neural networks are a general class of networks that has been explored by multiple groups \cite{tfn, kondor2018clebsch, 3dsteerable} and build on previous work on building equivariances into convolutional neural networks \cite{Worrall_2017_CVPR, KondorT18, CohenGW19}.

The success of convolutional neural networks at a variety of tasks is due to them having translation equivariance (e.g. a pattern can be identified in any location). Euclidean neural networks are a subset of convolutional neural networks where the filters are constrained to be equivariant to 3D rotations. To accomplish this, the filter functions are defined to be separable into a learned radial function and real spherical harmonics, $F_{lm}(\vec{r}) = R_{(l)}(|r|)Y_{lm}(\hat{r})$, analogous to the separable nature of the hydrogenic wavefunctions.

An additional consequence of Euclidean equivariance is that all ``tensors'' in a Euclidean neural network are geometric tensors and input and filter geometric tensors must be combined according to the rules of tensor algebra, using Clebsch-Gordan coefficients or Wigner 3j symbols (they are equivalent) to contract representation indices. We express these geometric tensors in an irreducible representation basis. The only convention in these networks is the choice of the basis for the irreducible representations of $O(3)$ which dictates the spherical harmonics and Wigner 3j symbols we use.

In our experiments, we use geometric tensors to express spatial functions; specifically the resulting coefficients from projecting a local point cloud onto spherical harmonics. We treat a local point cloud $S$ around a chosen origin as a set of $\delta$ functions and evaluate the spherical harmonics at those corresponding angles (up to some maximum $L$). Then, we weigh the spherical harmonic projection of each point by its radial distance from the origin

% MODIFY TO ACCOMMODATE BASIS ADJUSTMENT
\begin{align}
    f_S(\vec{x}) = \sum_{\vec{r} \in S} f_{\vec{r}} (\vec{x})
    = \sum_{\vec{r} \in S} \sum_{J=0}^L \underbrace{\|\vec r\| Y_J(\frac{\vec r}{\|\vec r\|})}_{F_J} \cdot Y_J(\vec{x})
\end{align}
% This procedure is common step in calculating rotation invariant descriptors of local atomic environments, such as smooth overlap of atomic positions (SOAP) kernels \cite{PhysRevB.87.184115}.
The coefficients of this projection $\sum_{\vec{r} \in S} F_J$ form a geometric tensor in the irreducible basis. We interpret the magnitude of the function on the sphere as a radial distance from the origin. We additionally re-scale this signal to account for finite basis effects by ensuring the max of the function corresponds to the original radial distance ($f_{\vec r}(\frac{\vec r}{\| \vec r \|}) = \|\vec r\|$).
%, Figure~\ref{fig:tetra}.

% \begin{figure}
%   \caption{A tetrahedron depicted as a point set (left) and as a projection onto spherical harmonic functions centered on the red point for $L \le 5$ (right). The function magnitude and surface coloring is plotted to be proportional to radial distance. The broadening of the lobes and the small lobes toward the center are artifacts due to the truncation of the spherical harmonic basis. \label{fig:tetra}}
%   \centering
%     \includegraphics[width=0.5\textwidth]{spherical_harmonics_geometry_features.png}
% \end{figure}

First, we prove that symmetry equivariant functions obey ``Curie's principle''. For a group $G$, vector space $V$, and a representation $D$, the symmetry group of $x \in V$ is defined as
\begin{equation}
    \label{eqn:symm_def}
    \text{Sym}(x) = \{g \in G : D(g)x = x\}.
\end{equation}

Let $f: V_1 \rightarrow V_2$ be equivariant to group $G$. ``Curie's principle'' can be articulated as
\begin{equation}
    \label{eqn:symm_out}
    \text{Sym(x)} \subseteq \text{Sym}(f(x)).
\end{equation}
Proof: For $g \in \text{Sym}(x)$ (i.e. $D_1(g)x = x$),
\begin{align}
    D_2(g) f(x) &= f(D_1(g) x) = f(x) \\
    &\Rightarrow g \in \text{Sym}(f(x)) \qquad\blacksquare
\end{align}

According to Eqn.~\ref{eqn:symm_out}, since Euclidean neural networks are equivariant to Euclidean symmetry, the symmetry of the output can only be of equal or higher symmetry than the input. This implies that the network will also preserve any subgroup of Euclidean symmetry, e.g. point groups and space groups.

\begin{figure}
  \centering
    \includegraphics[width=0.5\textwidth]{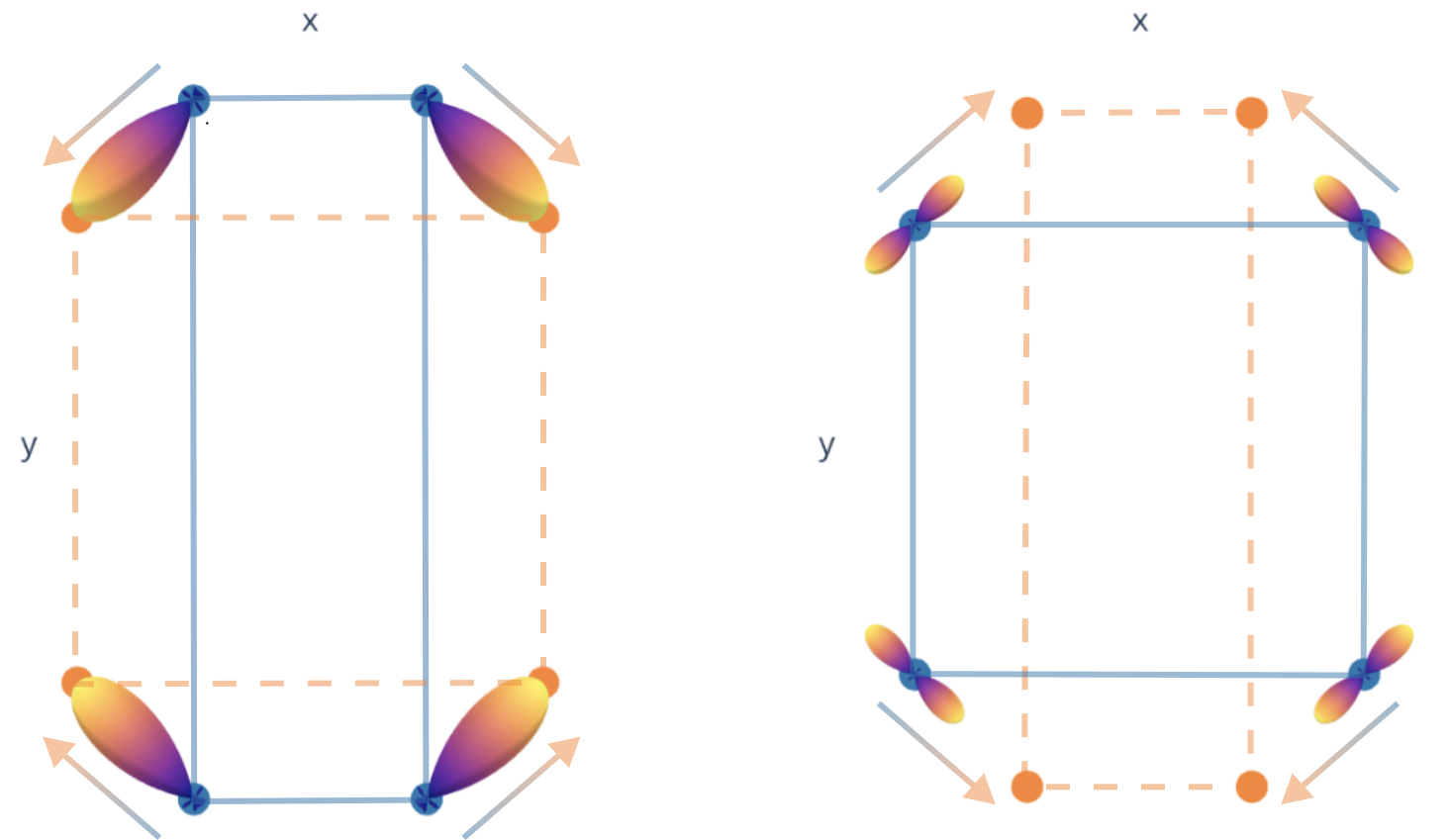}
    \caption{Left - The predicted displacement signal for a network trained to deform the rectangle into the square. Right - The predicted displacement signal for the network trained to deform the square into the rectangle. Solid lines outline the input shape and dashed lines outline the desired shape. Arrows point from input geometry to desired geometry.}
    \label{fig:task1and2}
\end{figure}

To demonstrate this, we train Euclidean neural networks to deform two arrangements of points in the $xy$ plane into one another, one with four points at the vertices of a square, and another with four points at the vertices of a rectangle, shown as blue and orange points in Figure~\ref{fig:task1and2}.

To conduct our experiments, we use the \texttt{e3nn} framework \cite{e3nn} for 3D Euclidean equivariant neural networks in this work written with PyTorch \texttt{pytorch} \cite{pytorch}. The \texttt{jupyter} \cite{jupyter} notebooks used for running the experiments and creating the figures for this Letter are made available at Ref. \cite{code}.

We train each network to match the spherical harmonic projection of the desired displacement vector i.e. final point location. As we will show, this representation is helpful for identifying degeneracies when they arise.

First, we train a Euclidean neural network to deform the rectangle into the square. This network is able to accomplish this quickly and accurately. Second, we train another Euclidean neural network to deform the square into the rectangle. No matter the amount of training, this network cannot accurately perform the desired task. 

In Figure~\ref{fig:task1and2}, we show output of the trained networks for both cases. On the right, we see that the model trained to deform the square into the rectangle is producing symmetric spherical harmonic signals each with two maxima. Due to being rotation equivariant, the network cannot distinguish distorting the square to form a rectangle aligned along the $x$ axis from a rectangle along the $y$ axis. The model automatically weighs symmetrically degenerate possibilities equally. By Eqn.~\ref{eqn:symm_out}, the output of the network has to have equal or higher symmetry than the input.

We emphasize here that the network does not ``know'' the symmetry of the inputs; the network predicts a degenerate answer simply because it is constrained to be equivariant. This is analogous to how physical systems operate and why physical systems exhibit ``Curie's priniciple''.

Having a dataset where the ``inputs'' are higher symmetry that the ``outputs'' implies there is missing data -- an asymmetry waiting to be discovered. 
In the context of phase transitions as described by Landau theory \cite{Landau:1937obd}, symmetry-breaking factors are called order parameters. To update its weights, a neural network is required to be differentiable, such that gradients of the loss can be taken with respect to every parameter in the model. This technique can be extended to the input; we use this approach to recover symmetry-breaking order parameters.
% Because order parameters are tensor quantities, prior to symmetry equivariant methods, it has been cumbersome and unclear how to articulate order parameters in machine learning tasks.

To prove that this is possible, we must prove that the gradients of a $G$-invariant scalar loss (such as the MSE loss) evaluated on the output of a $G$-equivariant neural network $f(x)$ and ground truth data $y_{true}$, e.g. $\partial (f(x) - y_{true})^2 / \partial{x}$, can have lower symmetry than the input.

The symmetry of the combined inputs to the invariant loss function is equal to or higher than the intersection of the symmetries of the predicted and ground truth outputs

\begin{align}
    \label{eqn:symm_combo}
    \text{Sym}(x) \cap \text{Sym}(y) \subseteq \text{Sym}(\alpha x + \beta y) \\
    \forall x,y \in V, \alpha, \beta \in \mathbb{R}.\nonumber
\end{align}
Proof: For $g \in \text{Sym}(x) \cap \text{Sym}(y)$,
\begin{equation}
    D(g) (\alpha x + \beta y) = \alpha D(g) x + \beta D(g) y  = \alpha x + \beta y \qquad\blacksquare
\end{equation}
Furthermore, if $\mathcal{L}$ is a differentiable and invariant function $\mathcal{L}: V \rightarrow \mathbb{R}$, then $\nabla \mathcal{L}$ is equivariant to $G$ by the equivariance of differentiation
\begin{equation}
    \label{eqn:symm_grad}
    \nabla f(D(g)x) = [D(g)^{-T}] \nabla \mathcal{L}(x).
\end{equation}
Thus, if the symmetry of the ground truth output is lower than the input to the network, the gradients can have symmetry lower than the input, allowing for the use of gradients to update the input to the network to make the network input and output symmetrically compatible. This procedure can be used to find symmetry-breaking order parameters missing in the original data but implied by symmetry.

Now, we demonstrate the symmetry properties of Euclidean neural networks according to Eqns. \ref{eqn:symm_combo} and \ref{eqn:symm_grad} can be used to learn symmetry breaking order parameters to deform the square into the rectangle.

In this task, we allow for additional inputs for each point of irreps $1_e \oplus 1_o \oplus 2_e \oplus 2_o \oplus 3_e \oplus 3_o \oplus 4_e \oplus 4_o$ where the number denotes the irrep degree $L$ and the subscript denote even and odd parity, respectively (e.g. $1_o$ transforms as a vector and $1_e$ transforms a pseudovector). These irreps are initialized to be zero and we modify the training procedure. We require the input to be the same on each point, such that we learn a ``global'' order parameter. We also add an identical component-wise mean absolute error (MAE) loss on each $L > 0$ components of the input feature to encourage sparsity. We train the network in the coordinate frame that matches the conventions of point group tables.

\begin{figure}
  \caption{Input parameters (top row) and output signal (bottom row) for one of the square vertices at (from left to right) the start, middle, and end of the model and order parameter optimization. For simplicity, we only plot components with the same parity as the spherical harmonics ($2^2_e$ and $4^2_e$) as they require less familiarity with how parity behaves to interpret. The starting input parameter (on the left) is only a scalar of value 1 ($0^0_e = 1.$), hence it being a spherically symmetric signal. As the optimization procedure continues, the symmetry-breaking parameters become larger, gaining contribution from components other than $0^0_e$ and the model starts to be able to fit to the target output. When the loss converges, the input parameters have non-zero $2^2_e$, $3^{-2}_e$, $4^2_e$, and $5^{-2}_e$ components with other non-scalar components close to zero and the model is able to fit to the target output. \label{fig:evolution}}
  \centering
    \includegraphics[width=0.5\textwidth]{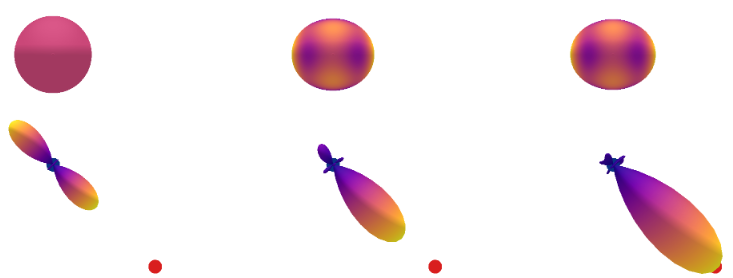}
\end{figure}

We first train the model normally until the loss no longer improves. Then we alternate between updating the parameters of the model and updating the input using gradients of the loss. As the loss converges, we find that the input for $L > 0$ consists of non-zero order parameters comprising only of $2^{2}_e$, $3^{-2}_e$, $4^{2}_e$, and $5^{-2}_e$, where the superscript denotes the order $m$ of the irrep where $-L \le m \le L$. See Fig. ~\ref{fig:evolution} for images of the evolution of the input and output signals during the model and order parameter optimization process. The order parameters distinguish the $x$ direction from the $y$ direction while maintaining the full symmetry of the rectangle. 

Our optimization returns four order parameters $2^{2}_e$, $3^{-2}_e$, $4^{2}_e$, and $5^{-2}_e$ because the gradients cannot break the degeneracy between these equally valid order parameters. To recover only e.g. the $2^{2}_e$ order parameter using Euclidean neural networks, we can do one of two things to break this degeneracy: limit the possible input order parameters e.g. $1_e \oplus 1_o \oplus 2_e \oplus 2_o$ or add a loss that penalizes higher degree $L$ order parameters. Thus, Euclidean neural networks can recover both the most general order parameters (including degeneracies) and more constrained orders parameters e.g. by using a custom loss function.

To arrive at this conclusion from the perspective of a conventional symmetry analysis: First, the symmetry of the square and rectangle must be identified as point group $D_{4h}$ and point group $D_{2h}$, respectively. Second, the lost symmetries need to be enumerated; going from the square to the rectangle, 8 symmetry operations are lost -- two four-fold axes, two two-fold axes, two improper four-fold axes, and two mirror planes. Then, the character table for the point group $D_{4h}$ is used to find which direct sum of irreps break these symmetries. In this case, there is one 1-dimensional irreducible representation of $D_{4h}$ that breaks all these symmetries, $B_{1g}$. The character table additionally lists that irrep $B_{1g}$ has a basis function of $x^2-y^2$ (i.e. $2^2_e$) in the coordinate system with $z$ being along the highest symmetry axis and $x$ and $y$ aligned with two of the mirror planes. Character tables only typically report basis functions up to $L \le 3$, so the higher order irreps $3^{-2}_e$, $4^{2}_e$, and $5^{-2}_e$ are not listed, but one can confirm with simple calculations that they transforms as $B_{1g}$. This conventional approach becomes more involved for objects with more complicated symmetry. In such cases, it is standard practice to employ computer algorithms to find e.g. relevant isotropy subgroups. However, many databases and tools for performing conventional symmetry analyses are not open source, making them difficult to incorporate into specific efforts.

\begin{figure}
  \caption{Perovskite crystal structure with chemical formula of the form ABX$_3$ and parent symmetry of $Pm\bar{3}m$ (221). Octahedra can tilt in alternating patterns. This increases the size of the unit cell need to describe the crystal structure. The larger unit cell directions are given in terms of the parent unit cell directions. The tilting of rotation axes for the $Pnma$ (62) structure is made of a 3D ``checkerboard'' of alternating rotations in the plane perpendicular to $a + b$ and a 2D ``checkboard'' of alternating rotations in the $ab$ plane along the $c$. The tilting of rotation axes for the $Imma$ (74) structure does not possess any alternating tilting in the $ab$ plane direction around the $c$ direction. The structure in space group $Pnma$ (62) correspond to Glazer notation $a^+ b^- b^-$ and Ref.~\cite{Howard1998} notation $(a000bb)$. The structure in space group $Imma$ (74) corresponds to Glazer notation $a^0 b^- b^-$ and Ref.~\cite{Howard1998} notation $(0000bb)$. \label{fig:perov}}
  \centering
    \includegraphics[width=0.5\textwidth]{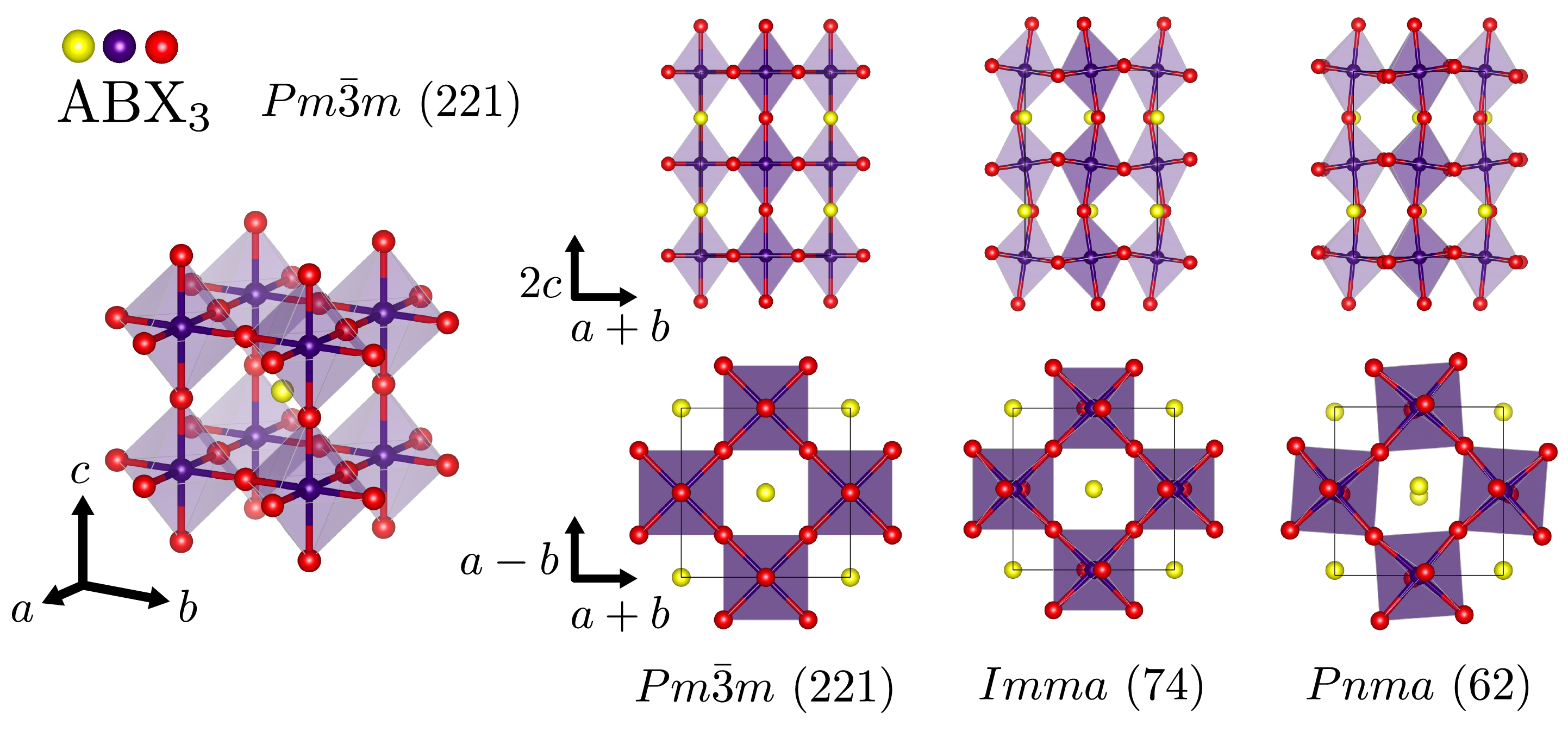}
\end{figure}

Now, we demonstrate this method on a more complicated example and use it to find symmetrically intermediate structures. Perovskite crystal structures are composed of octahedrally coordinated transition metal sites on a cubic lattice where the octahedra share corner vertices (Figure~\ref{fig:perov}). Perovskites display a wealth of exotic electronic and magnetic phenomena, the onset of which is often accompanied by a structural distortion of the parent structure in space group $Pm\bar{3}m$ (221) caused by the softening of phonon modes or the onset of magnetic orders \cite{Benedek2013}.

The octahedra in perovskites can distort in a variety of ways, one of which is by developing tilting patterns, commonly classified using Glazer notation introduced in Ref.~\cite{Glazer1972}.
Using the same procedure as the previous section, we recover the order parameters for two perovskites structures with different octahedral tilting.
We use periodic boundary conditions to treat the crystal as an infinite periodic solid.

For this demonstration we compare the parent structure in $Pm\bar{3}m$ (221) to the structure in the subgroup $Pnma$ (62).
We use the same training procedure as above except for the following: we only apply order parameters to the $B$ sites and we allow each $B$ site to have its own order parameter.
We also add a penalty that increases with the $L$ of the candidate order parameter.

From training, the model is able to recover that each $B$ site has a nontrivial pseudovector order parameter with equal magnitude; this can be intuitively interpreted as different rotation axes with equal rotation angle for each $B$ site. The pattern of rotation axes and corresponding octahedral tilting is described and shown in Figure~\ref{fig:perov}. If we look at the character table for $Pm\bar{3}m$ we can confirm that this pattern of pseudovectors matches the irreps $M^{3+} \oplus R^{4+}$, the irreps recovered in Ref.~\cite{Howard1998}. In contrast to conventional symmetry analysis, our method provides a more clear geometric interpretation of these order parameters as rotation axes. Additionally, the same model can be used to determine the form of the order parameter and build a model that can predict the amplitude of this distortion e.g. based on composition and the parent structure.

We can also learn to produce output that is symmetrically intermediate between input and ground truth output by restricting learnable order parameters. If we train an identical model, but constrain the pseudovector order parameter to be zero along the $c$ direction and non-adjacent B sites to have identical order parameters, we recover an intermediate structure in the space group $Imma$ (74) described and shown in Figure~\ref{fig:perov}.

In contrast to conventional symmetry analysis which requires classifying the symmetry of given systems, we perform symmetry analyses with Euclidean neural networks by learning equivariant mappings. This allows us to gain symmetry insights using standard neural network training procedures. Our methods do not rely on any tabulated information and can be directly applied to tensor fields of arbitrary complexity.

Symmetry equivariant neural networks act as ``symmetry compilers'': they can only fit data that is symmetrically compatible and can be used to help find symmetry-breaking order parameters necessary for compatibility. The properties proven in this Letter generalize to any symmetry-equivariant network and are relevant to any branch of physics using symmetry-aware machine learning models to create surrogate or generative models of physical systems. The same procedures demonstrated in this Letter can be used to find order parameters of other physical systems, e.g. missing environmental parameters of an experimental setup (such as anisotropy in the magnetic field of an accelerator magnet) or identifying other symmetry-implied information unbeknownst to the researcher.

% NEED TO UPDATE THIS
\section{Acknowledgements}

\begin{acknowledgments}
T.E.S. thanks Sean Lubner, Josh Rackers, Sin\'{e}ad Griffin, Robert Littlejohn, James Sethian, Tamara Kolda, Frank No\'{e}, Bert de Jong, and Christopher Sutton for helpful discussions. T.E.S. and M.G. were supported by the Laboratory Directed Research and Development Program of Lawrence Berkeley National Laboratory and B.K.M. was supported by CAMERA both under U.S. Department of Energy Contract No. DE-AC02-05CH11231.
\end{acknowledgments}

\bibliography{apssamp}% Produces the bibliography via BibTeX.

\end{document}